\documentclass[conference]{IEEEtran}
\IEEEoverridecommandlockouts
\usepackage[letterpaper, left=0.625in, right=0.625in, bottom=1in, top=0.75in]{geometry}
\usepackage{graphicx}
\usepackage[ruled,vlined,linesnumbered]{algorithm2e}
\usepackage{bm}
\usepackage{amssymb,mathtools,amsthm,amsfonts,amsmath}
\usepackage{cite}
\usepackage[dvipsnames]{xcolor}
\usepackage{url}
\usepackage{hyperref}
\usepackage{upgreek}
\usepackage{yfonts}
\usepackage{cleveref}
\usepackage{multirow}
\usepackage{balance}
\usepackage[subrefformat=parens,labelformat=parens,caption=false]{subfig}
\usepackage{enumitem}
\SetKwComment{Comment}{}{}

\newcommand{\vbf}{\mathbf{v}}
\newcommand{\wbf}{\mathbf{w}}
\newcommand{\xbf}{\mathbf{x}}

\newcommand{\Ibf}{\mathbf{I}}

\newcommand{\Sbf}{\mathbf{S}}

\newcommand{\Bcal}{{\mathcal{B}}}
\newcommand{\Ccal}{{\mathcal{C}}}

\newcommand{\Jcal}{{\mathcal{J}}}

\newcommand{\Ncal}{{\mathcal{N}}}

\newcommand{\Scal}{{\mathcal{S}}}

\newcommand{\Ucal}{{\mathcal{U}}}

\newcommand{\Ebb}{\mathbb{E}}
\newcommand{\Rbb}{\mathbb{R}}

\newcommand{\Pbb}{\mathbb{P}}

\DeclareMathAlphabet\mathbfcal{OMS}{cmsy}{b}{n}

\newtheoremstyle{ehsan}
  {}
  {}
  {}
  {\parindent}
  {\itshape}
  {:}
  { }
  {\thmname{#1}\thmnumber{ #2}\textit{\thmnote{ (#3)}}}

\theoremstyle{ehsan}

\renewcommand{\IEEEQED}{\hfill \IEEEQEDopen}

\title{Partial Model Sharing Improves Byzantine Resilience in Federated Conformal Prediction}

\author{\IEEEauthorblockN{{Ehsan Lari}\textsuperscript{1}, \text{Reza Arablouei}\textsuperscript{2}, \text{Stefan Werner}\textsuperscript{1,3}}
\textsuperscript{1}\text{Department of Electronic Systems, Norwegian University of Science and Technology, Trondheim, Norway}
\\ \textsuperscript{2}\text{CSIRO's Data61, Pullenvale QLD 4069, Australia}
\\ \textsuperscript{3}\text{Department of Information and Communications Engineering, Aalto University, Finland}
\thanks{This work was partially supported by the Research Council of Norway and the Research Council of Finland (Grant 354523).}
}
\makeatletter
\newcommand{\linebreakand}{%
  \end{@IEEEauthorhalign}
  \hfill\mbox{}\par
  \mbox{}\hfill\begin{@IEEEauthorhalign}
}

\begin{document}

\makeatother

\maketitle

\begin{abstract}

We propose a Byzantine-resilient federated conformal prediction (FCP) method that leverages partial model sharing, where only a subset of model parameters is exchanged each round. Unlike existing robust FCP approaches that primarily harden the calibration stage, our method protects both the federated training and conformal calibration phases. During training, partial sharing inherently restricts the attack surface and attenuates poisoned updates while reducing communication. During calibration, clients compress their non-conformity scores into histogram-based characterization vectors, enabling the server to detect Byzantine clients via distance-based maliciousness scores and to estimate the conformal quantile using only benign contributors. Experiments across diverse Byzantine attack scenarios show that the proposed method achieves closer-to-nominal coverage with substantially tighter prediction intervals than standard FCP, establishing a robust and communication-efficient approach to federated uncertainty quantification.

\end{abstract}

\section{Introduction} \label{sec:intro}

Federated learning (FL)~\cite{mcmahan2017communication, smith2017federated, bonawitz2019towards, kairouz2021advances, lari2023ssp, lari2023apsipa, lari2025admm} enables a network of devices, such as smartphones, media streamers, and Internet of Things (IoT) devices, to collaboratively train a global model without sharing raw data. This decentralized nature makes FL attractive for privacy-preserving applications. However, its reliance on open participation also exposes it to adversarial clients who can manipulate updates or submit misleading information. Among these adversaries, Byzantine clients are especially harmful because they participate in the learning protocol and can directly influence the outcome by behaving arbitrarily or maliciously, e.g., by transmitting corrupted gradients, injecting inconsistent model weights, or executing denial-of-service attacks that undermine the global model~\cite{guerraoui2018hidden, 9857216, chen2017distributed, fang2020local}. Therefore, ensuring robustness against such threats is essential for the practical deployment of FL.

Uncertainty quantification (UQ) has become a critical component of high-stakes machine learning (ML) applications in healthcare, finance, and defense~\cite{vovk2005algorithmic}. Unlike single-point predictions, UQ methods provide calibrated sets of plausible outcomes with statistical coverage guarantees, offering both accuracy and reliability. Conformal prediction (CP)~\cite{vovk2005algorithmic, balasubramanian2014conformal, lei2018distribution, romano2020classification} is a principled UQ framework that produces distribution-free prediction sets under minimal assumptions of exchangeability. To extend these benefits to distributed settings, federated conformal prediction (FCP) has been introduced~\cite{lu2023federated, humbert2023one}, enabling uncertainty-aware learning across multiple data sources. However, FCP directly inherits the vulnerabilities of FL: Byzantine clients can compromise both the training and calibration phases, for instance, by mislabeling high-risk financial profiles as safe, thereby corrupting global predictions in sensitive domains.

Recently, Rob-FCP~\cite{kang2024certifiably} has been proposed to mitigate the impact of Byzantine clients in FCP. Rob-FCP computes local non-conformity scores, sketches them into characterization vectors, and detects malicious clients by measuring distances between these vectors. Clients deemed highly malicious are then excluded from the calibration process. While effective in protecting calibration, Rob-FCP does not address training-time poisoning, as adversaries can already degrade the learned model during federated training, which in turn propagates into the conformal intervals. Moreover, Rob-FCP introduces additional client--server communication for robust calibration, since clients must transmit characterization statistics in addition to standard training messages, which can be costly at scale.


In this paper, we propose a Byzantine-resilient FCP framework built around \emph{partial sharing online federated learning (PSO-Fed)}~\cite{9746228, 9933811, lari2024icassp, lari2024analyzing}. The key idea is to couple communication-efficient partial parameter exchange with robust conformal calibration so that resilience is achieved \emph{across both stages} of the FCP pipeline. During federated training, clients exchange only a subset of model coordinates, which both reduces client--server communication and inherently attenuates the effect of poisoned updates by limiting the adversary’s influence on the aggregated model. During calibration, clients compress their local non-conformity scores into histogram-based characterization vectors and the server assigns distance-based maliciousness scores to detect and exclude Byzantine contributors prior to estimating the conformal quantile. Experiments under various attack models show that the proposed approach is more resilient to Byzantine attacks compared to standard FCP and Rob-FCP.

\section{Preliminaries} \label{sec:pre}

In this section, we first introduce the system model and review the PSO-Fed algorithm~\cite{9746228, 9933811, lari2024icassp, lari2024analyzing}. We then describe the Byzantine attack model in the context of FL and analyze its potential impact on PSO-Fed. Finally, we outline how CP operates in an FL setup.

\subsection{System Model}

We consider a federated network consisting of $K$ clients that can communicate with a central server. At each time instance $n$, client $k$ observes a data pair $\xbf_{k, n} \in \Rbb^D$ and $y_{k, n} \in \Rbb$, which follow the model
\begin{align} \label{eq1}
    y_{k, n} = \wbf^{\star\intercal} \xbf_{k, n} + \nu_{k, n},
\end{align}
where $\wbf^\star \in \Rbb^D$ is the unknown parameter vector to be estimated collaboratively and $\nu_{k, n}$ denotes observation noise.
The global learning task is to estimate $\wbf$ by minimizing the expected mean-squared error (MSE) across all clients:
\begin{align} \label{eq2}
    \Jcal(\wbf) = \frac{1}{K} \sum\limits_{k=1}^{K} \Jcal_k(\wbf),
\end{align}
where the local objective function at client $k$ is
\begin{align} \label{eq3}
    \Jcal_k(\wbf) & = \Ebb \left[|y_{k, n} - \wbf^\intercal \xbf_{k, n}|^2 \right].
\end{align}
The primary goal of FL is to obtain the optimal minimizer $\arg\min_{\wbf} \Jcal(\wbf)$ through decentralized collaboration between clients and the central server without exchanging raw data.

\subsection{PSO-Fed Algorithm}

In our prior work~\cite{lari2024icassp, lari2024analyzing}, we have studied PSO-Fed in detail. Here, we briefly summarize its operation. Unlike conventional FL, where the full parameter vector is exchanged between the server and clients, PSO-Fed communicates only a subset of parameters at each iteration. Specifically, the server sends a fraction of the global model estimate to each client, while each client transmits only a fraction of its local model estimate to the server. The diagonal selection matrix $\Sbf_{k, n} \in \Rbb^{D \times D}$ with $M$ ones identifies the model parameters communicated between client $k$ and the server at iteration $n$. Therefore, the PSO-Fed recursions minimizing \eqref{eq2} are given by
\begin{align*} 
\epsilon_{k, n} & = y_{k, n} - \left[ \Sbf_{k, n-1} \wbf_{n-1} +  \left( \Ibf_D -\Sbf_{k, n-1} \right) \wbf_{k, n-1} \right]^\intercal \xbf_{k, n}  \\ 
\wbf_{k, n} & = \Sbf_{k, n-1} \wbf_{n-1} +  \left( \Ibf_D - \Sbf_{k, n-1} \right) \wbf_{k, n-1} + \mu \xbf_{k, n} \epsilon_{k, n}  \\ 
\wbf_{n} & = \frac{1}{|\Scal_n|} \sum_{k \in \Scal_n} \left[ \Sbf_{k, n}  {\wbf}_{k, n} + \left( \Ibf_D - \Sbf_{k, n} \right) \wbf_{n-1} \right], 
\end{align*}
where $\wbf_{k, n}$ is the local model estimate at client $k$ and iteration $n$, $\wbf_n$ is the global model estimate at iteration $n$, $\Ibf_D$ is the $D \times D$ identity matrix, $\mu$ is the stepsize controlling convergence speed and steady-state performance, $\Scal_n$ is the set of selected clients at iteration $n$, and $|\Scal_n|$ is the cardinality of $\Scal_n$.

\subsection{Byzantine Attack during Training}

Let $\Scal_B$ denote the set of Byzantine clients in the network, and let $\beta_k$ be an indicator variable such that $\beta_k = 1$ if client $k \in \Scal_B$ and $\beta_k = 0$ otherwise. The server is assumed to know $|\Scal_B|$.
At each iteration, a Byzantine client may corrupt its transmitted (partial) update with probability $p_a$. Specifically, we model the corruption as $\wbf_{k, n} + \tau_{k, n} \boldsymbol{\delta}_{k, n}$, where $\tau_{k, n}$ is a Bernoulli random variable with $\Pr(\tau_{k,n}=1)=p_a$, and $\boldsymbol\delta_{k,n} \sim \Ncal ( \mathbf{0}, \sigma_B^2 \Ibf_D )$ is zero-mean white Gaussian noise~\cite{7563348}. Accordingly, the server receives $\Sbf_{k,n}\wbf_{k, n} + \beta_k \tau_{k, n} \Sbf_{k,n}\boldsymbol \delta_{k, n}$ instead of the benign partial update $\Sbf_{k,n}\wbf_{k, n}$.

\subsection{Federated Conformal Prediction}

FCP for regression generates predictive intervals that achieve a user-specified confidence level (e.g., $90\%$) under the assumption of data exchangeability. CP calculates non-conformity scores, typically residuals on a calibration set, and uses their quantiles to form prediction intervals with rigorous marginal coverage guarantees. Importantly, these guarantees hold regardless of the underlying model's accuracy or complexity, making CP a distribution-free method for UQ.

In the standard CP setting, a calibration dataset $\{ ( \xbf_j, y_j ) \}_{j=1}^N$ and a trained model $\hat{\wbf}$ are required. The non-conformity score for each calibration point is
\begin{align}
    r_j = |y_j - \hat{\wbf}^\intercal \xbf_j|, \quad j = 1, \dots, N.
\end{align}
Let $q_{1 - \alpha}$ denote the $1 - \alpha$ quantile of these scores:
\begin{align}
    q_{1 - \alpha} = \mathrm{Quantile}_{1 - \alpha} \left( \{ r_1, \dots, r_N \} \right).
\end{align}
Then, the two-sided $1 - \alpha$ prediction interval for a new input $\xbf$ is
\begin{align}
    \Ccal (\xbf) = \left[ \hat{\wbf}^\intercal \xbf - q_{1 - \alpha},\; \hat{\wbf}^\intercal \xbf + q_{1 - \alpha} \right].
\end{align}
The confidence parameter $\alpha\in (0, 1)$ controls interval width: smaller $\alpha$ (higher confidence) yields wider intervals, while larger $\alpha$ produces narrower but less conservative ones. By construction, the marginal coverage satisfies
\begin{align}
    \Pbb \left( y \in \Ccal ( \xbf) \right) \geq 1 - \alpha.
\end{align}

In FL, after training a global model $\hat{\wbf}$, each client $k$ holds a local calibration set $\{ ( \xbf_{k, j}, y_{k, j} ) \}_{j=1}^{N_k}$ and computes local non-conformity scores
\begin{align}
    r_{k, j} = |y_{k, j} - \hat{\wbf}^\intercal \xbf_{k, j}|,\ j = 1, \dots, N_k.
\end{align}
Privacy constraints prevent clients from sharing raw calibration data or even individual scores. The FCP algorithm~\cite{lu2023federated} addresses this by employing secure aggregation or sketching techniques (e.g., CountSketch), which allow the server to approximate the global quantile as
\begin{align}
\hat{q}_{1 - \alpha} \approx \mathrm{Quantile}_{1 - \alpha} \left( \bigcup_{k=1}^K \{ r_{k, j} \}_{j=1}^{N_k} \right).
\label{eq:quantile}
\end{align}
This enables the server to construct CP-style intervals that aim to achieve near-nominal global coverage while preserving the privacy of individual clients’ calibration sets. Under mild conditions (e.g., non-pathological score distributions), the approximation incurs only negligible degradation in empirical coverage.

\subsection{Byzantine Attack during Calibration}

\begin{figure*}[t!]
\centering
\subfloat[Efficiency attack\label{fig:conv-vs-nrobustADMM}]{{\includegraphics[width=.325\textwidth]{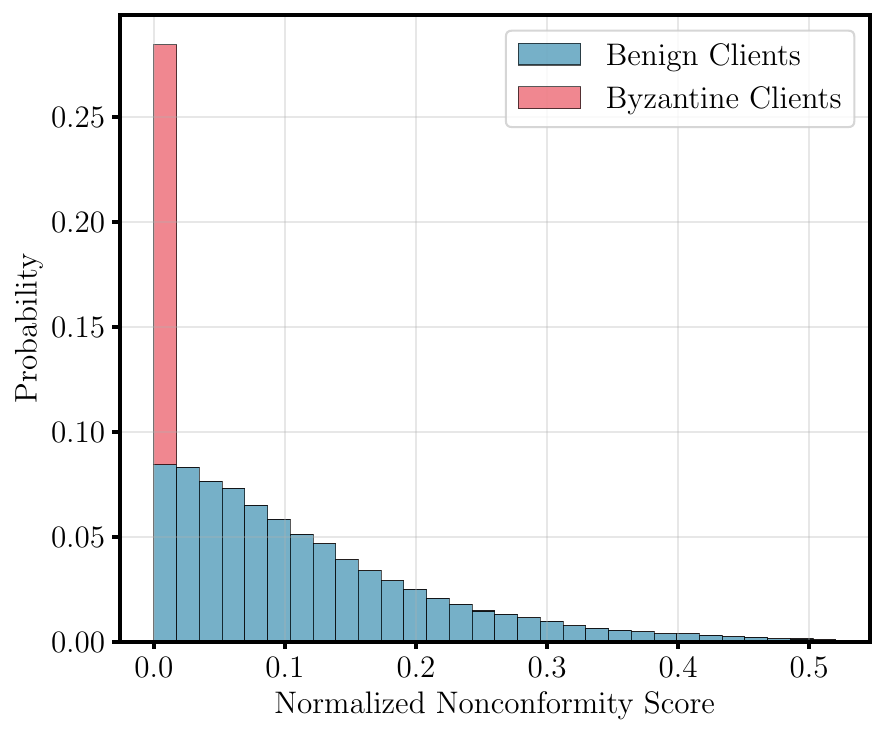}}} \hspace{1mm}
\subfloat[Coverage attack\label{fig:conv-vs-nrobustADMM_diff}]{{\includegraphics[width=.325\textwidth]{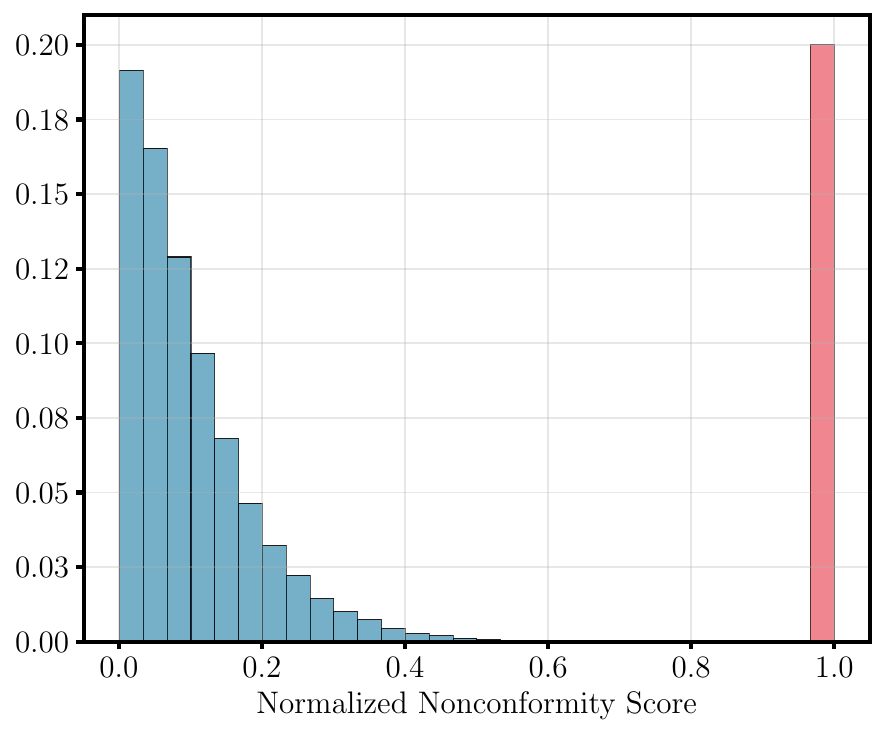}}} \hspace{1mm}
\subfloat[Random attack\label{fig:prop_nrADMM_diff}]{{\includegraphics[width=.325\textwidth]{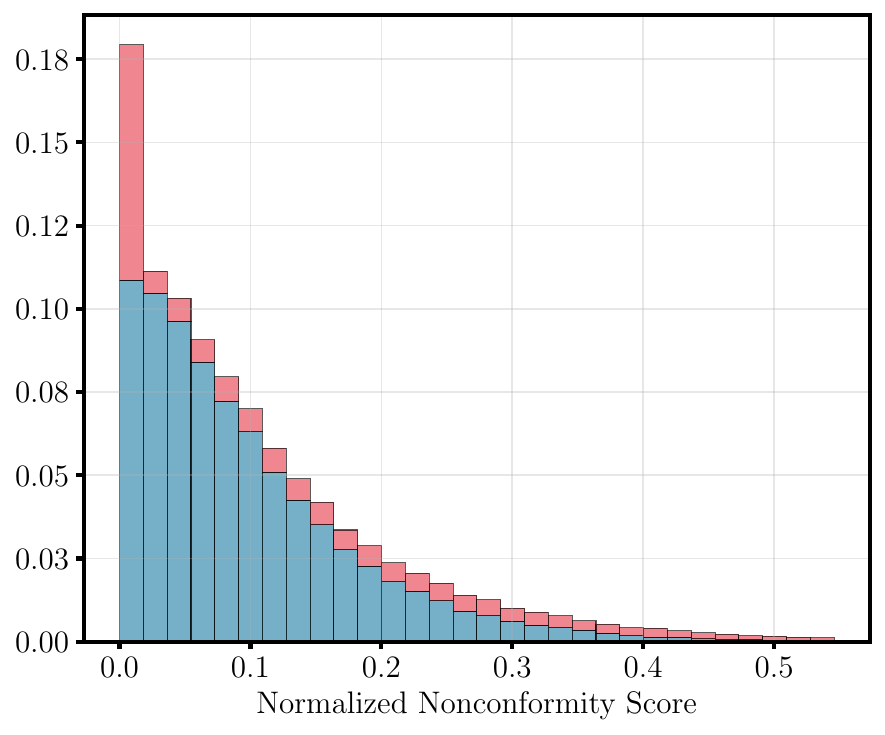}}}
\caption{Example histograms illustrating the effect of different Byzantine attacks during the calibration phase: (a) efficiency attack, (b) coverage attack, and (c) random attack.}
\label{fig:DiffAttacks}
\end{figure*}

During the calibration phase, Byzantine clients in $\Scal_B$ may submit arbitrary or adversarially crafted non-conformity scores in place of their true values, with the aim of skewing the global quantile estimate used for interval construction. Such manipulations can distort prediction intervals, making them either too wide (overly conservative) or too narrow (undercovering), thus undermining the coverage guarantees of CP. 

In this work, we consider three representative types of Byzantine attacks, whose effects are illustrated in Fig.~\ref{fig:DiffAttacks}: 
\begin{itemize}[leftmargin=*]
\item \emph{Efficiency attack}, clients report abnormally small scores, deflating $q_{1-\alpha}$ leading to undercoverage.
\item \emph{Coverage attack}, clients report artificially large scores, inflating $q_{1 - \alpha}$ and producing excessively wide prediction intervals. 
\item \emph{Random attack}, clients submit random scores, injecting noise into the quantile estimate.
\end{itemize}

\section{Byzantine-Resilient FCP via Partial Sharing} \label{sec:method}

In this section, we present our approach for mitigating the impact of Byzantine attacks during both the training and calibration phases of FCP.

\subsection{Mitigating Byzantine Attack during Training Phase}

As shown in~\cite{lari2024icassp, lari2024analyzing}, PSO-Fed reduces communication by allowing each client to share only a fraction of its model parameters with the server. Beyond communication savings, partial sharing has also been demonstrated to enhance robustness against model-poisoning attacks. In particular, we have proven that PSO-Fed maintains both mean and mean-square convergence even in the presence of Byzantine clients that inject noise into their updates. In this work, we leverage the partial sharing mechanism to mitigate the effect of Byzantine attacks during the training phase of FCP.

\subsection{Mitigating Byzantine Attacks during Calibration Phase}

To defend against Byzantine attacks in the calibration phase, each benign client computes non-conformity scores $\{ r_{k,j} \}_{j=1}^{N_k}$ on its local calibration set and summarizes them into a histogram-based characterization vector $\vbf_k \in \Delta^H$ defined as
\begin{align}
    v_{k, h} = \frac{1}{N_k} \sum_{j=1}^{N_k} \mathbf{1}\left[ a_{h-1} \le r_{k,j} < a_h \right],
\end{align}
where the score range is normalized to $[0,1]$ and partitioned into $H$ bins with boundaries $0 = a_0 < a_1 < \cdots < a_H = 1$. Each entry $v_{k,h}$ then denotes the empirical probability of local scores falling into bin $h$. Here,
\begin{align}
\Delta^H \triangleq \left\{ \vbf_k \in \mathbb{R}^H \;\middle|\; v_{k,h} \geq 0,\; \sum_{h=1}^H v_{k,h} = 1 \right\}
\end{align}
denotes the $H$-dimensional probability simplex.
The characterization vectors are transmitted to the server, while Byzantine clients may submit arbitrary or adversarial vectors (see Fig.~\ref{fig:DiffAttacks}). 

This histogram representation offers several advantages: it preserves the distributional structure of local non-conformity scores, facilitates reliable comparisons across clients, improves privacy by avoiding raw score transmission, and reduces communication overhead by compressing local information. Benign clients typically yield similar vectors, whereas Byzantine clients tend to deviate, enabling detection. 

To identify adversaries, we compute pairwise $\ell_2$ distances between characterization vectors as
\begin{align}
    d_{k, k^{\prime}} = \left\lVert \vbf_k - \vbf_{k^{\prime}} \right\rVert_2, \quad \forall\, k, k^{\prime} \in \{1, \cdots, K\}.
\end{align}
We then assign each client $k$ a maliciousness score $m_k$, defined as the average distance to its $K_b - 1$ nearest neighbors as
\begin{align} \label{mal_score}
    m_k = \frac{1}{K_b - 1} \sum_{k^{\prime} \in \mathrm{NN}(k, K_b-1)} d_{k,k^{\prime}},
\end{align}
where $K_b \triangleq K-|\Scal_B|$ denotes the number of benign clients. We classify the $K_b$ clients with the lowest maliciousness scores as benign, forming the set $\Bcal$. We then estimate quantile $\hat{q}_{1 - \alpha}$ only on $\Bcal$ to ensure robust federated conformal calibration. By adopting this strategy, our approach inherits the robustness guarantees of Rob-FCP while additionally benefiting from communication efficiency, yielding valid and reliable prediction intervals even in the presence of Byzantine clients.

\begin{figure*}[t!]
\centering
\subfloat[Efficiency attack\label{fig:ZeroAttack}]{\includegraphics[width=.3333\textwidth]{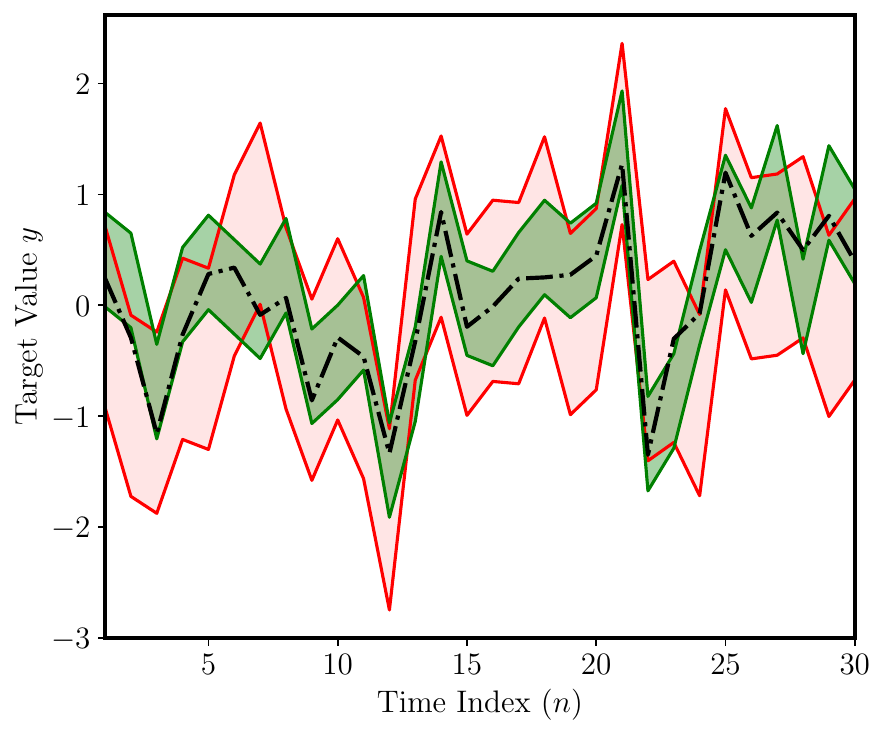}}
\subfloat[Coverage attack\label{fig:OneAttack}]{\includegraphics[width=.3333\textwidth]{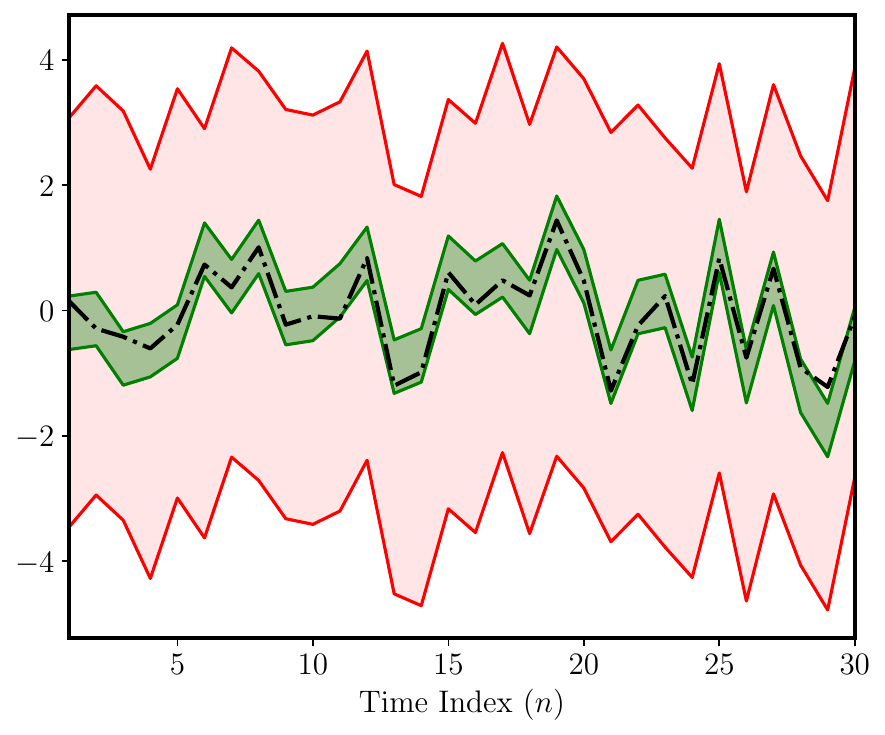}}
\subfloat[Random attack\label{fig:RandomAttack}]{\includegraphics[width=.3333\textwidth]{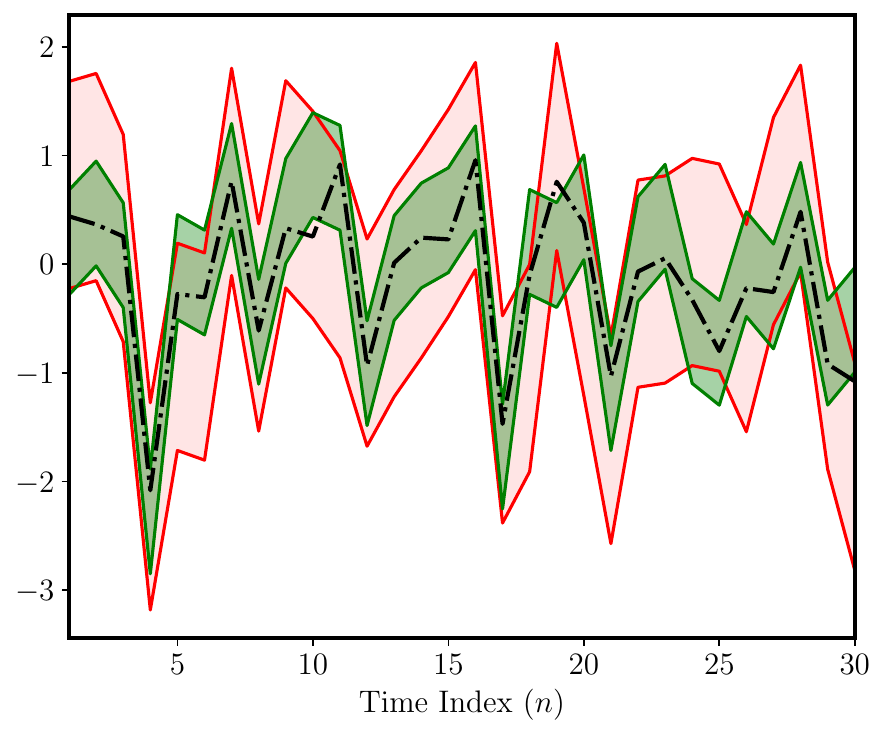}}
\caption{Illustrative prediction intervals under (a) efficiency, (b) coverage, and (c) random attacks. The true target values are shown as dashed lines, while prediction intervals from FCP and our partial sharing approach with $M/D = 0.3$ are shown in \textcolor{BrickRed}{red} and \textcolor{ForestGreen}{green}, respectively.}
\label{fig:FLBLUEvsonesh}
\end{figure*}
\begin{figure}[t!]
    \centering
    \includegraphics[width=0.83\columnwidth]{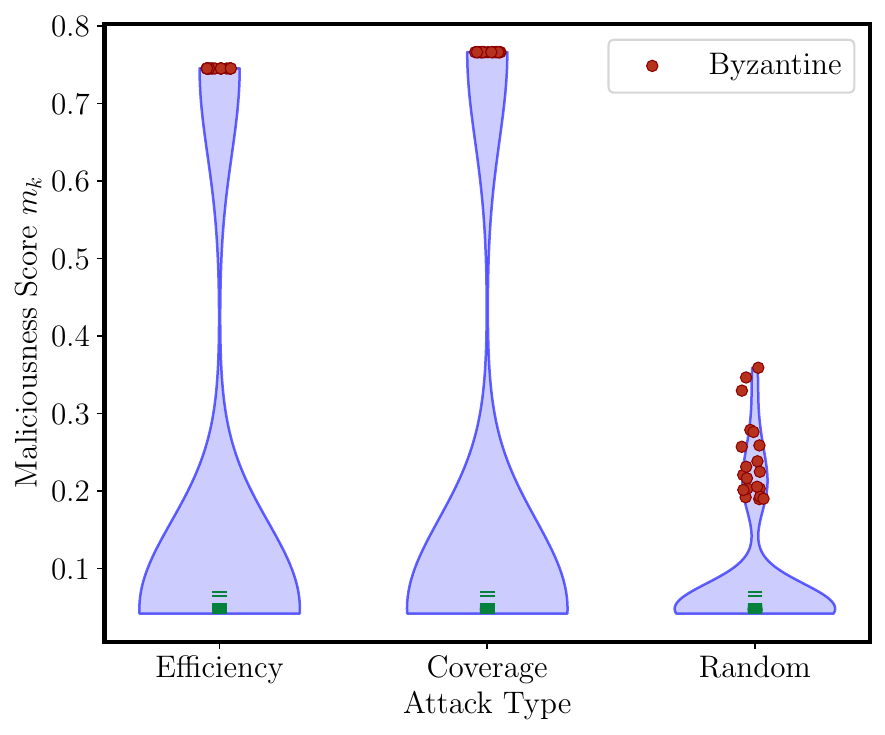}
    \caption{Violin plots illustrating the distribution of maliciousness scores $m_k$ for different types of Byzantine attacks.}
    \label{fig:Boxplot}
\end{figure}

\section{Simulation Results} \label{sec:simulations}

In this section, we present numerical experiments to evaluate the performance of partial model sharing in FCP under various Byzantine attack scenarios. We consider a federated network of $K = 100$ clients. At each iteration, the server randomly selects $|\Scal_n| = 10$ clients to participate in training. Each client $k$ holds non-IID data consisting of an input vector $\xbf_{k,n} \in \Rbb^{D}$ and a corresponding response $y_{k,n}$, related through~\eqref{eq1} with a unit-energy model parameter $\wbf^\star=\frac{1}{\sqrt{D}}[1,\cdots,1]^\intercal \in \Rbb^{D}$. 
We set the model dimension to $D=50$ and communicate a random subset of $M = 15$ parameters ($M/D = 0.3$) between each client and the server at each iteration. The entries of $\xbf_{k,n}$ are drawn from zero-mean Gaussian distributions with client-specific variances $\varsigma_k^2 \sim \Ucal(0.2, 1.2)$, while the observation noise is Gaussian with variance $\sigma_{\nu_k}^2 \sim \Ucal(0.005, 0.025)$.

In our experiments, we consider $|\Scal_B| = 20$ Byzantine clients. During the training phase, these clients employ an attack probability of $p_a = 0.25$ and inject Gaussian noise with variance $\sigma_B^2 = 0.1$ into their updates. 
During the calibration phase, as illustrated in Fig.~\ref{fig:DiffAttacks}, the same Byzantine clients report adversarial non-conformity scores according to three attack scenarios: (i) \emph{all zeros}, (ii) \emph{all ones}, and (iii) \emph{uniform random scores} drawn from the uniform distribution $\Ucal(0.8, 1)$. 
We fix the size of the characterization vectors to $H = 100$, and set the target coverage to $90\%$, i.e., $\alpha = 0.1$ in \eqref{eq:quantile}.

In Fig.~\ref{fig:Boxplot}, we demonstrate the effectiveness of our approach in handling Byzantine behavior by depicting the maliciousness scores of all clients under efficiency, coverage, and random attacks (see Fig.~\ref{fig:DiffAttacks}). In each case, Byzantine clients exhibit significantly higher scores and are clearly identifiable as outliers, enabling their reliable exclusion from the calibration process. Note that without prior knowledge about the number of Byzantine clients, another efficient way to detect outliers is to use the \emph{median absolute deviation (MAD)} rule~\cite{leys2013mad}.

\begin{figure}
    \centering
    \subfloat[Coverage]{\includegraphics[width=0.925\columnwidth]{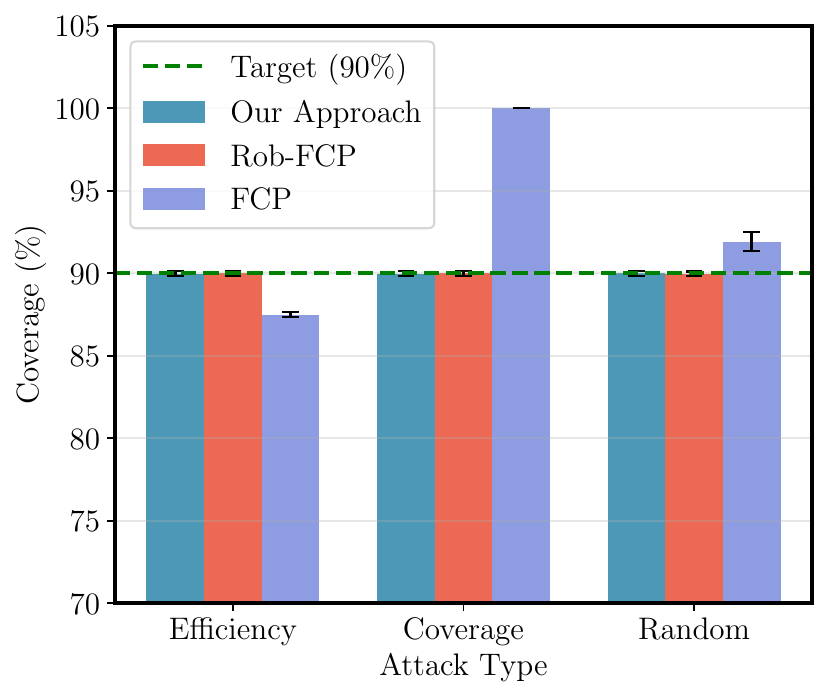}}\\
    \centering
    \subfloat[Interval width]{\includegraphics[width=0.895\columnwidth]{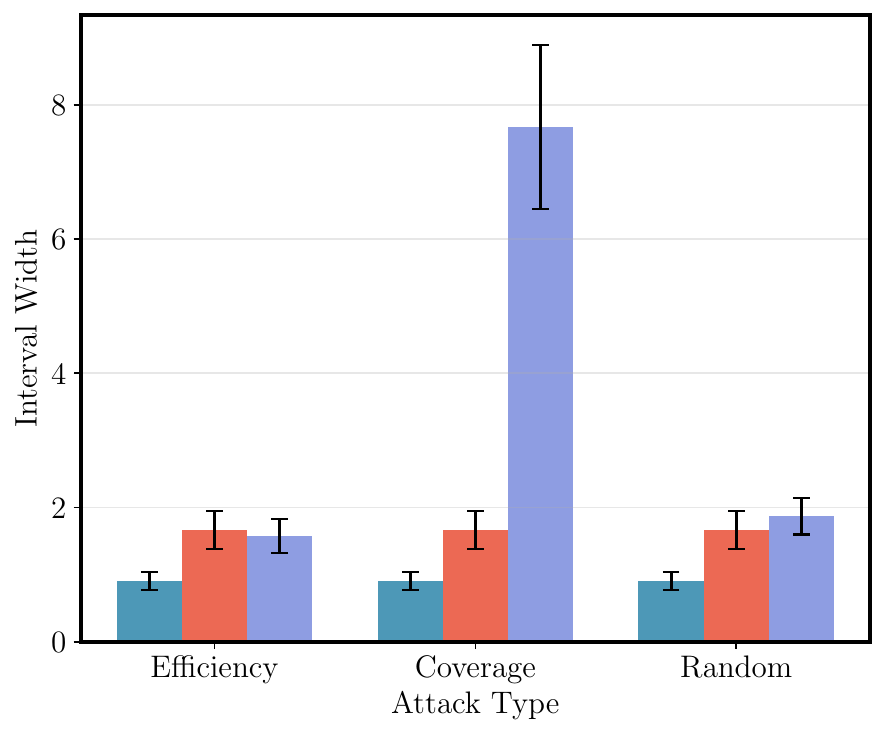}}
    \caption{Comparison of FCP and Rob-FCP with our approach under Byzantine attacks ($D=50$, 20\% Byzantine clients).
    }
    \label{fig:method-comparison}
\end{figure}

We present the marginal coverage and average interval width for both FCP with and without partial model sharing under the identical Byzantine attack scenarios in Fig.~\ref{fig:FLBLUEvsonesh} and Fig.~\ref{fig:method-comparison}.
We conduct training, calibration, and testing using $1{,}000$ synthetic data samples per client, with the first $30$ samples shown in Fig.~\ref{fig:FLBLUEvsonesh} for clear visualization.

In Fig.~\ref{fig:method-comparison}, we compare FCP and Rob-FCP with our approach under Byzantine attacks that affect both the training and calibration phases. FCP, which performs no Byzantine filtering, is vulnerable: under the coverage attack, it yields inflated coverage ($\approx93\%$) together with overly wide intervals (about $2.2\times$ that of ours). Rob-FCP mitigates calibration-phase manipulation by filtering Byzantine clients using the maliciousness scores $m_k$ (cf.\ \eqref{mal_score}), and therefore maintains near-nominal coverage ($\approx90\%$). However, because Rob-FCP does not mitigate training-phase attacks, its learned model exhibits higher MSE and consequently wider intervals. Our approach addresses both phases: partial sharing attenuates Byzantine influence during training (reducing MSE), while calibration-stage filtering preserves robust quantile estimation. As a result, ours produces intervals that are $1.8\times$ tighter than Rob-FCP and $2.2\times$ tighter than FCP.

Overall, incorporating PSO-Fed into Rob-FCP yields two key benefits: (i) \emph{resilience to adversarial behavior} during both training and calibration of CP, and (ii) \emph{enhanced communication efficiency} through random client participation and partial model parameter sharing. 

\section{Conclusion and Future Work} \label{sec:conclusion}

We developed a Byzantine-resilient CP framework by integrating PSO-Fed's partial sharing mechanism into Rob-FCP. 
During training, clients exchange only selected model coordinates, which attenuates the impact of poisoned updates while reducing communication. During calibration, clients summarize local non-conformity scores via histogram-based characterization vectors and the server assigns maliciousness scores to identify and exclude Byzantine clients.
Simulation results show that our approach attains closer-to-nominal coverage with significantly tighter prediction intervals compared to both standard FCP and Rob-FCP across a range of Byzantine attack scenarios. 
A promising direction for future work is to establish theoretical guarantees that quantify how partial sharing affects robustness and coverage, further strengthening the foundations of trustworthy FL.

\balance
\bibliographystyle{IEEEtran}
\bibliography{refs} 

\end{document}